\def\year{2021}\relax
\documentclass[letterpaper]{article} % DO NOT CHANGE THIS
\usepackage{aaai21-mitre}  % DO NOT CHANGE THIS
\usepackage{times}  % DO NOT CHANGE THIS
\usepackage{helvet} % DO NOT CHANGE THIS
\usepackage{courier}  % DO NOT CHANGE THIS
\usepackage[hyphens]{url}  % DO NOT CHANGE THIS
\usepackage{graphicx} % DO NOT CHANGE THIS
\usepackage{natbib}  % DO NOT CHANGE THIS AND DO NOT ADD ANY OPTIONS TO IT
\usepackage{caption} % DO NOT CHANGE THIS AND DO NOT ADD ANY OPTIONS TO IT
\usepackage{tabularx}

\title{Adversarial Attack Attribution:\\Discovering Attributable Signals in Adversarial ML Attacks}
\author{
Marissa Dotter,
Sherry Xie,
Keith Manville,
Josh Harguess,
Colin Busho,
Mikel Rodriguez
\\
}

\affiliations{
    The MITRE Corporation \\
    \{mdotter, stxie, kmanville, jharguess, cbusho, mikel\}@mitre.org

}

\begin{document}

\maketitle
% %%%%%%%%% TITLE
% \title{Beyond Defense: Discovering the Attribution of an Individual Adversarial Attack}

% \author{Marissa Dotter\\
% MITRE\\
% {\tt\small mdotter@mitre.org}
% \and
% Keith Manville\\
% {\tt\small kmanville@mitre.org}

% \and
% Josh Harguess\\
% {\tt\small jharguess@mitre.org}

% \and
% Colin Busho\\
% {\tt\small cbusho@mitre.org}

% \and
% Mikel Rodriguez\\
% {\tt\small mikel@mitre.org}
% }
% \maketitle
% %\thispagestyle{empty}

%%%%%%%%% ABSTRACT
\begin{abstract}
Machine Learning (ML) models are known to be vulnerable to adversarial inputs and researchers have demonstrated that even production systems, such as self-driving cars and ML-as-a-service offerings, are susceptible. These systems represent a target for bad actors. Their disruption can cause real physical and economic harm. When attacks on production ML systems occur, the ability to attribute the attack to the responsible threat group is a critical step in formulating a response and holding the attackers accountable. We pose the following question: can adversarially perturbed inputs be attributed to the particular methods used to generate the attack? In other words, is there a way to find a signal in these attacks that exposes the attack algorithm, model architecture, or hyperparameters used in the attack? We introduce the concept of adversarial attack attribution and create a simple supervised learning experimental framework to examine the feasibility of discovering attributable signals in adversarial attacks. We find that it is possible to differentiate attacks generated with different attack algorithms, models, and hyperparameters on both the CIFAR-10 and MNIST datasets.

\end{abstract}

%%%%%%%%% BODY TEXT
\section{Introduction}

\begin{figure}[h!]
\centering
\includegraphics[width=8.4cm]{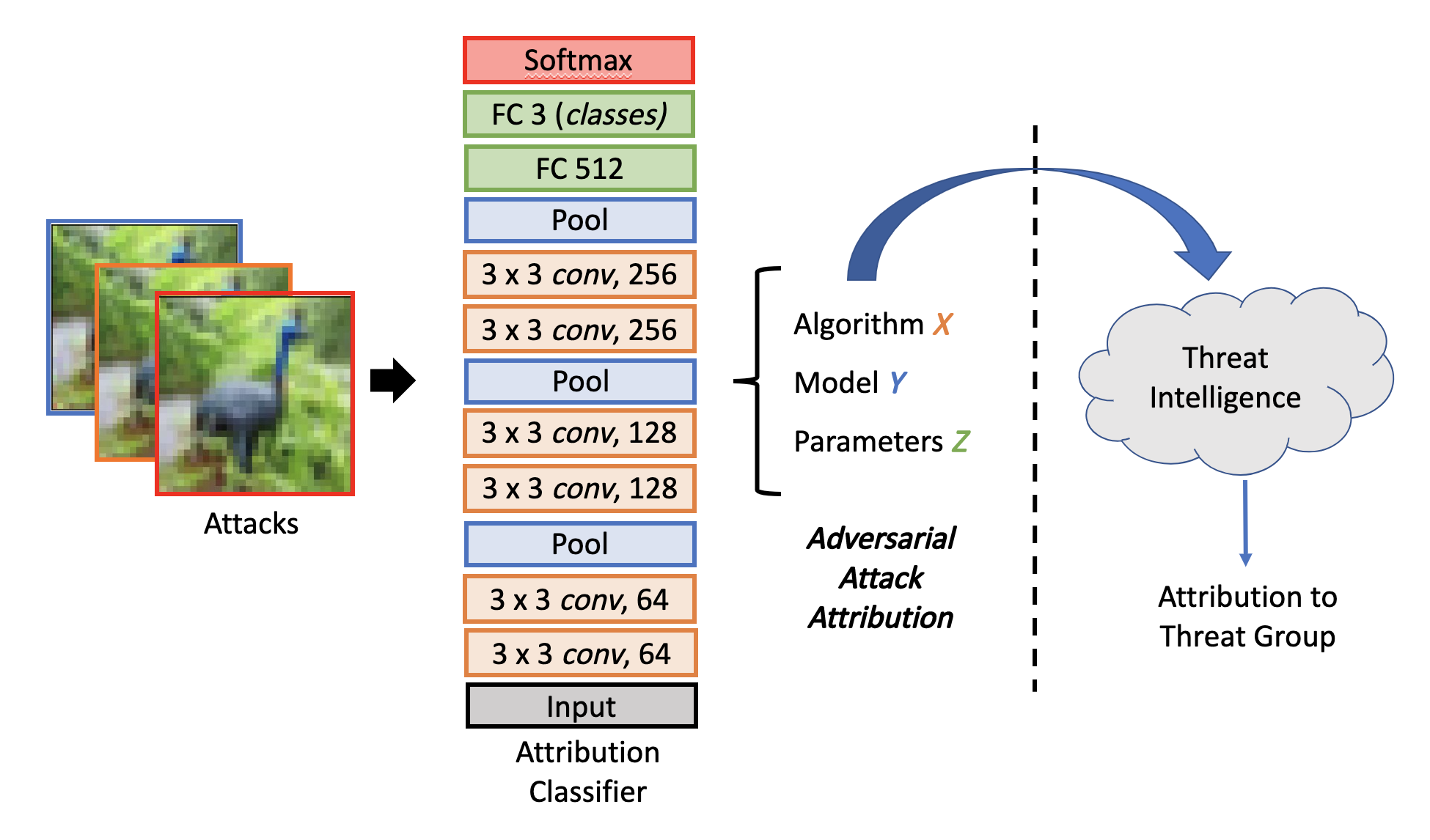}
\caption{Adversarial Attack Attribution Overview}
\label{fig:overview_attribution}
\end{figure}

As we collectively begin to use machine learning (ML) in safety critical systems and as components in large commercial systems, potential attacks on these systems become much more than an intellectual curiosity.
An adversary attacking a ML system could result in loss of life, disruption of service and economic damages, or theft of valuable intellectual property.
Researchers have demonstrated proof-of-concept attacks on several commercial systems, including Tesla's autopilot \cite{tencent2019tesla}, Amazon Alexa's speech recognition \cite{li2019advmusic}, and Microsoft Azure \cite{advmlthreatmatrix}.
It is only a matter of time before adversaries leverage these techniques to cause real harm.
Attribution can potentially deter cyber attacks or hold attackers accountable by leading to political sanctions or legal proceedings \cite{EU2, EU}.
In the cybersecurity world, two of the key indicators that enable cyber attribution are tradecraft and malware \cite{odni}.
Security analysts have identified that threat groups follow certain Tactics, Techniques and Procedures (TTPs) and the attacks they use may leave behind particular identifying signals.
TTPs and signatures can be used to help attribute an attack to a particular group.  With that goal in mind, Figure \ref{fig:overview_attribution} presents an overview of adversarial attack attribution.
Attribution is a larger concept that plays a key role alongside intent, infrastructure, and other cyber indicators in revealing tradecraft.
Identifying the types of adversarial attacks used could provide vital cyber threat information related to these indicators. Further analysis of these indicators could aid in the attribution of the attacks and hold bad actors accountable.

\begin{figure}
\centering
\includegraphics[width=8.3cm]{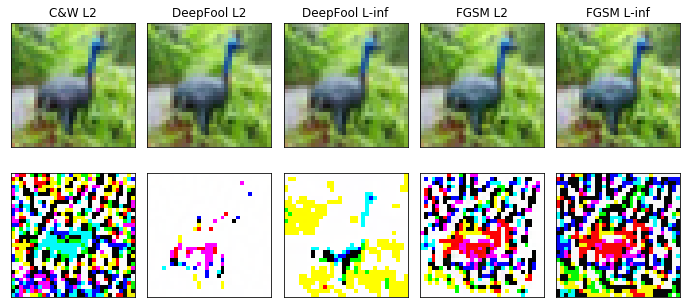}
\caption{An image from un-targeted attacks (top row) and the underlying perturbations (bottom row) on a CIFAR-10 image using the AlexNet CNN to create the attacks.}
\label{fig:attacks}
\end{figure}

Recent work has shown that modern ML approaches, in particular deep neural networks, suffer from being fooled by inputs that have been perturbed, or ``attacked.'' %These adversarial examples represent one class of attack and have grown in both popularity and complexity.
%and as we learn more about the internal functions of neural networks, the more diverse the attack algorithms have become. 
In response, algorithms that can detect and/or defend against certain types of adversarial attacks have become a large subspace of deep learning research. 
%An aspect that has only recently started to be discussed within these two sub-sections is adversarial attack attribution. 
Adversarial attack attribution attempts to further understand adversarial attacks by uncovering distinguishable signals between different types of attack algorithms. Its goal is to find some underlying signal in the perturbations as shown in Figure~\ref{fig:attacks}, that aids in understanding where distinctions between attacks can be found: 
%from the trained model, the attack algorithm hyperparameters, the framework used to create the algorithms, etc. 
from the attack algorithm, model architecture, or hyperparameters used in the attack. 
The focus of this work is on attacks that happen at inference time, however we expect that the concept of adversarial attack attribution will be extended to other types of attacks.
%such as those that occur during the learning process.
% while we are tracking that attacks can happen at learning time or inference time we are god using on a specific set of attacks (integrity at inference time). 
%We expect that the concept of adversarial attack attribution will be extended to other types of attacks. 
While research into this area is nascent, 
the Defense Advanced Research Projects Agency (DARPA), under the newly formed, Reverse Engineering of Deceptions (RED) effort \cite{darpa2020red}, is also interested in developing “techniques that automatically reverse engineer the tool-chains behind attacks such as multimedia falsification, adversarial ML attacks, or other information deception attacks.”

Adversarial attack attribution research may help us understand the techniques used by bad actors in practice. The ability to recover the process used to generate an attack could lend insights leading to better defenses. 
%This includes discovering the frameworks, architectures, and hyperparameters used to create attacks as well as the methods that adversaries use to introduce attacks to a system. 
This includes discovering the attack algorithm, model architecture, and hyperparameters used to create attacks as well as the methods that adversaries use to introduce attacks to a system.
With the continual growth of the machine learning field, many publicly available tools, frameworks, and even models can be downloaded and modified for adversarial needs \cite{papernot2016technical,rauber2017foolbox}. Attribution could further help determine the sophistication of the attacker, the resources they have available, and even characterize their methods from attacks that do not fit within these publicly available tools.

%http://www.activeresponse.org/wp-content/uploads/2013/07/diamond.pdf
%https://arxiv.org/pdf/1907.00374.pdf
%https://arxiv.org/pdf/1802.06430.pdf
%https://arxiv.org/abs/1911.00126

%The cyber security community has built a set of tools called ATT\&CK that organizes the methods used by malicious actors \cite{attack_phil}. The ML community has created a similar framework titled advmlthreatmatrix \cite{advmlthreatmatrix}. These tools demonstrate the real world need to defend against cyber attacks to include adversarial attacks. Attribution is part of a holistic defense

%-------------------------------------------------------------------------
% next paragraph is about this paper and the experimental methodology and research questions
% Though a real-world attack may not be this broad, by creating such a large experimental space we can begin to understand properties of attribution, where it succeeds in distinguishing attacks, and where it breaks down within our current approach.
This paper seeks to understand if adversarial ML attacks are attributable to the methods used to generate them, and thus aid in attribution to a particular threat group.
More specifically, from a data sample with an embedded attack, can we identify the attack algorithm, model, or attack hyperparameters used to generate the adversarial example?  Our experimental setup covers a broad-range of attack algorithms, models, and individual attack hyperparameters to attempt to answer that question.
By creating this experimental space we can begin to understand properties of attribution, where it succeeds in distinguishing attacks, and where it fails within our current approach. We can also work towards addressing the following questions:

\textbf{1. Are there underlying signals embedded within adversarial perturbations that can be used to classify the adversarial \textit{attack algorithm} that was used to create the adversarial dataset?} We present our experiments on multiple attack algorithms, created using the same model architecture, and optimized to have minimal perceptible perturbations, which maximize the adversarial accuracy of the data. We find that attack algorithm attribution is possible, and there does exist an underlying signal that particular algorithms leave behind that a classifier can discover.

\textbf{2. Are there underlying signals that can distinguish the \textit{model} that was used to create the adversarial dataset?} We train models with various architectures and use them to generate attacks. We find that particular models do leave behind underlying signals indicating that a classifier can uncover attribution of not only the attack algorithms, but also the architecture used to generate the attack. 

\textbf{3. Are there underlying signals that can be used to distinguish the individual attack algorithm \textit{hyperparameters} that were used to create the adversarial dataset?} We create variations of adversarial datasets with different attack algorithm hyperparameters and $L_p$ norm constraints to study the task of hyperparameter attribution. Our attribution framework can in some cases distinguish these hyperparameters, indicating that even subtle changes in the optimization of an attack algorithm can be discovered.    

The paper is organized as follows. We first summarize related work in adversarial machine learning as well as work in cyber attack attribution. Our methodology and experimental design for adversarial attack attribution follow. We then present the results of our experiments with a discussion of the results. Finally, we conclude the paper with our final thoughts and considerations for future work. 

% \begin{figure}[thb]
% \centering
% % \captionsetup{justification=centering}
% \begin{subfigure}[b]{\textwidth}
% \centering
% \title{The input to the Attribution CNN}
% \includegraphics[width=8.3cm]{figures/5attacks_cnn_originals.png}
% % \caption{}
% % \label{fig:sfig1}
% \end{subfigure}
% \begin{subfigure}[b]{\textwidth}
% \centering
% \title{The underlying perturbation we aim to attribute.}
% \includegraphics[width=8.3cm]{figures/5attacks_cnn_pertrubations.png}
% % \caption{}
% % \label{fig:sfig2}
% \end{subfigure}
% \caption{An image of each attack and their perturbations on an original CIFAR-10 image.}
% \label{fig:input_attacks}
% \end{figure}
%-------------------------------------------------------------------------
\section{Related Work}
\label{sec:prev}

This work relates quite closely to the literature of adversarial attacks and defenses in the context of real-world cyber attacks. We briefly introduce several well-known attacks \cite{akhtar2018threat} and defenses \cite{chakraborty2018adversarial}, touch on related work in fingerprinting GANs, and finally discuss cyber attribution. 
% This field is moving rapidly, so this is meant to be an introduction, not an exhaustive coverage, of the research in adversarial attacks and defenses. 

\textbf{Adversarial Attacks:} \cite{szegedy2013intriguing} is one of the earliest demonstrations of adversarial attacks on deep learning models. In that work, the authors added small perturbations to images that could fool deep learning models into highly confident misclassifications. In \cite{goodfellow2014explaining}, the authors present a simple, yet effective method of creating adversarial examples called the fast gradient sign method (FGSM) . 
% The Basic Iterative Method (BIM) \cite{kurakin2016adversarial} iteratively computes small steps, while adjusting the direction after each step, in contrast to one-step methods. There are also methods to build adversarial attacks by changing just a single pixel, such as the work by \cite{su2019one}. 
DeepFool \cite{moosavi2016deepfool} efficiently finds the minimal perturbation needed to fool a deep learning model. It does this by projecting the input onto the closest hyperplane and minimally perturbing the input until it is misclassified resulting in extremely small perturbations.
%Carlini \& Wagner \cite{carlini2017towards} introduced attacks to overcome defenses against the above attacks that successfully transfer from unsecured networks to secured making the attacks appropriate for black-box attacks. 
Carlini \& Wagner \cite{carlini2017towards} introduced a powerful optimization-based adversarial attack that has been shown to be effective even against defended models.

\textbf{Adversarial Defenses:} As research into attack types and variations grow, so do the defenses against those attacks. The earliest and most common approach to defense is adversarial training  \cite{goodfellow2014explaining}, which injects adversarial examples into the training set to increase model robustness. Using the fact that methods such as FGSM rely on the model’s gradient for the attack, gradient hiding defenses have been developed \cite{tramer2017ensemble}. Network distillation methods, which is a way to transfer knowledge from larger networks to smaller ones, \cite{papernot2017extending,papernot2016distillation} have also been developed to defend against adversarial attacks.
% The main idea of using distillation is the assumption that the secondary model will be robust to attacks that the original model may have been vulnerable to. Another way to view our attribution work presented here is as a defense to adversarial attacks by detecting potential adversarial samples. If you can accurately attribute an adversarial input to a particular framework, architecture, set of parameters, etc., then you can successfully defend against that input. 

\textbf{Adversarial Detection:} Much of the defensive literature has aimed to create models that are robust to adversarial attacks. However some works try to explicitly detect adversarial inputs \cite{pang2018robustdetection,grosse2017Ostatisticaldetection,xu2018featuresqueezing,Yang_Chen_Hsieh_Wang_Jordan_2020} so their predictions can be flagged as unreliable. Adversarial attribution can be seen as a fine-grained detection task. In this work, we operate under the assumption that an input has already been identified as adversarial.

\textbf{GAN Attribution:} In \cite{yu2019attributing}, the authors explore whether Generative Adversarial Networks have unique ``fingerprints'' that are discoverable in generated images. They find that images are attributable to a particular GAN. The authors also find that fingerprinting can help defend against attacks. Our work poses similar questions for adversarial examples. We explore attribution of not only the model, but the attack algorithm and its hyperparameters.

\textbf{Cybersecurity:} This work is directly related to the ongoing work of attribution of a particular cybersecurity attack or set of attacks to a bad actor or group of bad actors. 
An extensive summary of various techniques to perform attribution of computer attackers, as well as an introduction of a taxonomy of attribution techniques, is presented in \cite{wheeler2003techniques}.
In \cite{rid2015attributing}, the authors introduce the `Q Model' designed to explain, guide,
and improve the making of attribution of cyber attacks. 
An Enhanced Cyber Attack Attribution Framework is introduced in \cite{pitropakis2018enhanced} to detect and defend against Advanced Persistent Threats (APTs) and ultimately attribute the attack to malicious parties behind the campaign. 
The authors of \cite{guitton2013sophistication} argue that the level of `sophistication' of a cyber-attack is not necessarily an indication for attribution of an attack to a particular actor, as is often argued and assumed in cyber attack attribution. 
A framework for cyber attack attribution based on threat intelligence is introduced in \cite{qiang2016framework} which uses the `local advantage model' to analyze the process of cyber attacks.
In \cite{edwards2017strategic} the authors use a game-theoretic approach resulting in a `blame game' to analyze policy-relevant questions, including the attribution of cyber attacks.
In terms of real-world incidents within adversarial ML, MITRE, Microsoft and several other organizations have joined together to create the Adversarial ML Threat Matrix \cite{advmlthreatmatrix} that will allow security analysts to work with threat models in a similar way to how these analysts confront cybersecurity attacks.

\section{Methodology} 
\label{sec:methodology}

\begin{figure}
\centering
\includegraphics[width=8.4cm]{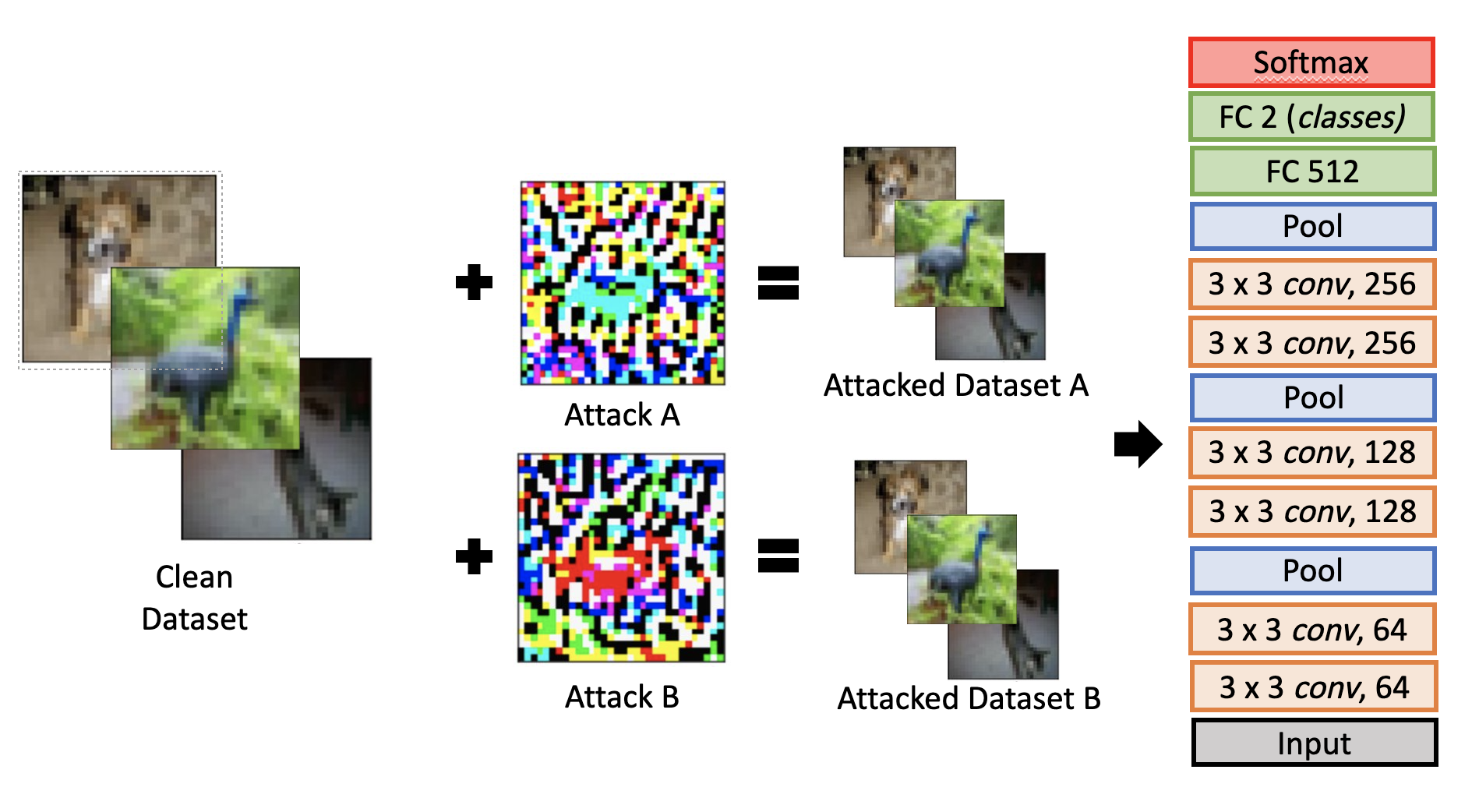}
\caption{Experimental Setup}
\label{fig:overview_train}
\end{figure}

In this work we pose the problem of adversarial attribution as a simple supervised learning problem as described by Figure~\ref{fig:overview_train}. We assume knowledge of the possible attack on the target ML system and the ability to generate attacked versions of the test data. We also assume that the data has already been identified as adversarial, therefore the identification of individual adversarial attacks is outside of the scope of this paper and considered future work. The focus of this work is solely on the attribution of these attacks. We say an attack is attributable if a classifier can be trained to distinguish between different attacks. We consider attacks that utilize different attack algorithms, models, and hyperparameters to generate datasets for our attribution experiments. A summary of the datasets created is in Table \ref{table:adv_datasets}. No comparisons to other algorithms are shown since we believe this is the first work of its kind at this time.

For our experiments we train baseline models AlexNet \cite{krizhevsky2012imagenet}, VGG16 \cite{simonyan2014very}, and ResNet50 \cite{he2016deep}) on the CIFAR-10 \cite{krizhevsky2009learning} and MNIST \cite{lecun2010mnist} datasets. 
% We additionally train a `2-layer Dense' network on the MNIST dataset for verification purposes, which is just a classification network with 2 dense layers followed by a softmax output. % may not need this
These models are trained with randomly initialized weights without pre-training or transfer-learned weights/layers, which limits extraneous variables presented to our attribution classifier. The three attacks used to create adversarial datasets from the CIFAR-10 and MNIST datasets were the Fast Gradient Sign Method (FGSM) \cite{goodfellow2014explaining}, DeepFool \cite{moosavi2016deepfool}, and Carlini \& Wagner (C\&W) \cite{carlini2017towards}. These attacks were chosen due to their simplicity and effectiveness, as well as to include optimized and non-optimized attacks for comparison. These attack algorithms were all implemented and modified from the Cleverhans library \cite{papernot2016technical}. An example of each adversarial attack and the underlying perturbation is shown in Figure~\ref{fig:attacks} on the CIFAR-10 dataset. We also consider attack hyperparameters including $L_2$ and $L_{\infty}$ norms. We optimized each attack algorithm as an un-targeted attack meaning that the attack algorithm aims to mis-classify each sample without targeting a specific class. The adversarial datasets were created combinatorially through parameter choices as shown in Table~\ref{table:adv_datasets}.

% These three architectures will serve as the basis by which we attempt to understand whether publicly available models leave a signal in the adversarial perturbations that is characteristic of that architecture.
% Hyperparameters for each attack and each dataset are chosen based upon recommended values from literature as well as their effectiveness in creating adversarial datasets. 
Only adversarial datasets that effectively fool the original architecture are retained for our attribution experiments. If the attack is not effective, it would not be of concern in a real-world setting. This helped ensure that our datasets were close to a real-world attack while remaining in this experimental framework. 

% moved to discussion
% Though not reported in these experiments, an interesting study between perceptibility and adversarial accuracy could be designed to investigate attribution of an attack. 

After creating the adversarial datasets across the different optimization parameters, we train a Convolutional Neural Network (CNN) with a standard cross-entropy loss for a binary or multi-class classification problem. The model we chose for these classification tasks had an
% standard 
architecture consisting of six convolutional layers and two fully connected layers followed by a softmax output as seen in Figure~\ref{fig:overview_train}.
% A CNN is an optimal choice for this task because they are notoriously good at learning complex distributions of images and it was our hypothesis that it would find some underlying signal in the adversarial datasets to make attribution possible. 
We use the standard CIFAR-10 and MNIST train/test splits, but with the attacked variants of the data for training and testing our attribution classifiers. We do not include an `unknown' class in these experiments, but note that such a class would be beneficial in future experiments to account for new models, hyperparameters, norms, or even attacks that are unseen or unknown.

%-------------------------------------------------------------------------
%\subsection{Experimental Design}

%We begin with the smallest subset of experiments when possible; those constrained within a single attack, before introducing new attacks. This allows understanding of the relationship between attacks more clearly when new attacks are introduced to the attribution classifier.

\begin{table}[h!]
% \caption{Wide single-column table in a twocolumn document.}
\centering

\begin{tabular}{p{0.14\linewidth}p{0.15\linewidth}p{0.408\linewidth}p{0.07\linewidth}} %p{0.24\linewidth}
\multicolumn{4}{c}{\textbf{Adversarial Attack Datasets}} \\
\hline
\textbf{Attack} & \textbf{Models} & \textbf{Hyperparameters} & \textbf{Norm}  \\
\hline
FGSM & AlexNet &  Epsilon: 1.0, 2.0, 5.0 & ${L_2}$  \\ 
 & VGG16  & Epsilon: 0.03, 0.1, 0.2 &   ${L_{\infty}}$ \\ 
 & ResNet50 & &  \\
\hline
DeepFool & AlexNet & Overshoot: 0.01, 0.1, 1.0 & ${L_2}$ \\ 
 & VGG16 & Overshoot: 0.01, 0.1, 1.0 & ${L_{\infty}}$ \\ 
 & ResNet50 & &  \\
\hline
C\&W & AlexNet & LR: 0.1, 0.2, 0.5 & ${L_2}$ \\ 
 & VGG16 & Conf.: 0.01, 0.1, 1.0 & \\ 
 & ResNet50 & &  \\
\hline
\end{tabular}
\caption{An overview of the adversarial attack algorithms, the models used to create each attack, and the hyperparameters the algorithms were optimized with. Through combinations of `Attack', `Model', `Hyperparameters', and `Norm' we create a diverse set of adversarial datasets to use throughout our experiments, i.e. FGSM ${L_2}$ Epsilon 1.0 Model: ResNet50 is a single adversarial dataset. Total datasets include: FGSM ${L_2}$: 9 datasets, FGSM ${L_{\infty}}$: 9 datasets, DeepFool ${L_2}$: 9 datasets, DeepFool ${L_{\infty}}$: 9 datasets, C\&W ${L_2}$: 27 datasets.}
\label{table:adv_datasets}
\end{table}

\section{Results}
\label{sec:results}

The results of our experimentation are organized by the three initial questions we sought to answer. All of our results are displayed as attribution accuracy: the fraction of attacked images that were correctly attributed by our classifier. In all experiments, each attacked dataset is the same size, making it easy to compare accuracy to random performance (random = $\frac{1}{\#~ datasets}$). We consider attribution successful if our classifier performs better than random by a statistically significant margin. Bold values of the overall attribution accuracy indicate successful attribution within this framework. Where results are broken down to per-class accuracy, accuracy is intentionally not bold for readability of the table. Our experimental results show attribution using the full adversarial dataset however, we note that experimentation through constraining the number of adversarial examples we train on could give further insight into attribution as it approximates a real-world scenario.

% Tables

% \subsection{CIFAR-10}

\subsection{Attack Algorithm Attribution}
The first question we sought to answer through these experiments was whether or not there was some underlying perturbation signal that could be distinguished from one attack algorithm to another. We display results for all three attack algorithms. Table~\ref{table:cifar_attack_attrib} and Table~\ref{table:mnist_attack_attrib} display the results of three-class attack algorithm attribution for CIFAR-10 and MNIST on a per model basis. We consider DeepFool and FGSM with different norms ($L_2$ and $L_\infty$) separately. We expand the $L_2$ case to a nine-class problem that considers  attack algorithm and model attribution simultaneously. Figure \ref{fig:cifar10_attacks} and Figure \ref{fig:mnist_attacks} are confusion matrices for CIFAR-10 and MNIST respectively.

As we can see for both the CIFAR-10 and MNIST datasets, attack algorithms are able to be distinguished with high confidence on the test set. 
%Even as we examine the per-class accuracy for each attack algorithm, we can see that for almost all classes, we achieve an accuracy higher than random chance, which indicates that no particular class is more robust in being attributed than another.
For example, as we examine the per-class accuracy for each attack algorithm (right three columns of Tables \ref{table:cifar_attack_attrib} and \ref{table:mnist_attack_attrib}), we can see that for almost all classes, we achieve an accuracy higher than random chance.

\begin{table*}[ht!]
% \caption{Wide single-column table in a twocolumn document.}
\centering
%\begin{tabular}{c{0.35\linewidth}c{0.15\linewidth}c{0.1\linewidth}c{0.1\linewidth}c{0.1\linewidth}c{0.1\linewidth}}
\begin{tabular}{cccccc}
\multicolumn{6}{c}{\textbf{CIFAR-10 Adversarial Attack Algorithm Attribution}} \\
\hline
\textbf{Attack} &  \textbf{Model} & \textbf{Attribution} & \multicolumn{3}{c}{\textbf{Per-Class Accuracy}} \\
\textbf{Algorithms} & & \textbf{Accuracy} & \textbf{C\&W} & \textbf{DeepFool} & \textbf{FGSM} \\
\hline
% (1) vs (2) & AlexNet & \textbf{0.55} & 0.79 & 0.30 & -- \\
%  & VGG16 & \textbf{0.53} & 0.36 & 0.70 & -- \\
%  & ResNet50 & \textbf{0.57} & 0.74 & 0.39 & -- \\
% \hline
% (2) vs (5) & AlexNet &  \textbf{0.83} & -- & 0.71 & 0.95  \\
%  & VGG16 & \textbf{0.71} & -- & 0.82 & 0.59 \\
%  & ResNet50 & \textbf{0.73} & -- & 0.86 & 0.60  \\
% \hline
% (3) vs (4) & AlexNet & \textbf{0.80} & -- & 0.70 & 0.90 \\
%  & VGG16 & \textbf{1.0} & -- & 1.0 & 1.0 \\
%  & ResNet50 & \textbf{0.85} & -- & 0.90 & 0.80 \\
% \hline
% (1) vs (4) & AlexNet & \textbf{0.95} & 0.94 & -- & 0.97 \\
% & VGG16 & \textbf{0.96} & 0.96 & -- & 0.97 \\
% & ResNet50 & \textbf{0.98} & 0.97  & -- & 0.98 \\
% \hline
% (1) vs (5) & AlexNet & \textbf{0.72} & 0.91 & -- & 0.52 \\
% & VGG16 & \textbf{0.66} & 0.46 & -- & 0.86 \\
% & ResNet50 & \textbf{0.74} & 0.79 & -- & 0.68 \\
% \hline
C\&W ${L_2}$ vs DeepFool ${L_2}$ vs FGSM ${L_2}$ & AlexNet & \textbf{0.70}  & 0.62 & 0.48 & 0.97 \\
& VGG16 & \textbf{0.70}  & 0.30 & 0.83 & 0.97 \\
& ResNet50 & \textbf{0.78} & 0.57 & 0.78  & 0.99 \\
\hline
C\&W ${L_2}$ vs DeepFool ${L_{\infty}}$ vs FGSM ${L_{\infty}}$ & AlexNet & \textbf{0.58} & 0.48 & 0.38  & 0.87 \\
& VGG16 & \textbf{0.50} & 0.60 & 0.10 & 0.79 \\
& ResNet50 & \textbf{0.56} & 0.16 & 0.75 & 0.77 \\
\hline
\end{tabular}
\caption{Reporting attribution accuracy for cross attack algorithms as per class accuracy per architecture (model) on CIFAR-10. The hyperparameters chosen for each attack maximize adversarial accuracy and minimize perceptibility and are described in Table~\ref{table:adv_datasets}. This table presents the overall accuracy of the attack algorithm attribution under `Attribution Accuracy' followed by the breakdown of per class accuracy.}
\label{table:cifar_attack_attrib}
\end{table*}

\begin{table*}[ht!]
% \caption{Wide single-column table in a twocolumn document.}
\centering
%\begin{tabular}{c{0.35\linewidth}c{0.15\linewidth}c{0.1\linewidth}c{0.1\linewidth}c{0.1\linewidth}c{0.1\linewidth}}
\begin{tabular}{cccccc}
\multicolumn{6}{c}{\textbf{MNIST Adversarial Attack Algorithm Attribution}} \\
\hline
\textbf{Attack} &  \textbf{Model} & \textbf{Attribution} & \multicolumn{3}{c}{\textbf{Per-Class Accuracy}} \\
\textbf{Algorithms} & & \textbf{Accuracy} & \textbf{C\&W} & \textbf{DeepFool} & \textbf{FGSM} \\
\hline
% (1) vs (2) & AlexNet & \textbf{0.83} & 0.97 & 0.71 & -- \\
%  & VGG16 & \textbf{0.92} & 0.97 & 0.88 & -- \\
%  & ResNet50 & \textbf{0.93} & 0.97 & 0.90 & -- \\
% \hline
% (1) vs (3) & AlexNet & \textbf{0.99} & 0.99 & 0.99 & -- \\
%  & VGG16 & \textbf{1.0} & 1.0 & 1.0 & -- \\
%  & ResNet50 & \textbf{0.99} & 0.99 & 0.99 & -- \\
% \hline
% (1) vs (4) & AlexNet & \textbf{0.82} & 0.72 & -- & 0.93 \\
% & VGG16 & \textbf{0.71} & 0.44 & -- & 0.98 \\
% & ResNet50 & \textbf{0.93} & 0.87  & -- & 1.0 \\
% \hline
% (1) vs (5) & AlexNet & \textbf{1.0} & 1.0 & -- & 1.0 \\
% & VGG16 & \textbf{0.99} & 0.99 & -- & 0.99 \\
% & ResNet50 & \textbf{1.0} & 1.0 & -- & 1.0 \\
% \hline
% (2) vs (4) & AlexNet & \textbf{0.59} & -- & 0.23 & 0.94 \\
% & VGG16 & \textbf{0.77} & -- & 0.87 & 0.69 \\
% & ResNet50 & \textbf{0.59} & -- & 0.87 & 0.32 \\
% \hline
% (2) vs (5) & AlexNet &  \textbf{0.99} & -- & 0.99 & 0.99  \\
%  & VGG16 & \textbf{1.0} & -- & 1.0 & 1.0 \\
%  & ResNet50 & \textbf{0.99} & -- & 0.99 & 0.99  \\
% \hline
% (3) vs (4) & AlexNet & \textbf{0.99} & -- & 0.99 & 0.99 \\
%  & VGG16 & \textbf{1.0} & -- & 1.0 & 1.0 \\
%  & ResNet50 & \textbf{0.99} & -- & 0.99 & 0.99 \\
% \hline
% (3) vs (5) & AlexNet & \textbf{1.0} & -- & 1.0 & 1.0 \\
% & VGG16 & \textbf{1.0} & -- & 1.0 & 1.0 \\
% & ResNet50 & \textbf{0.99} & -- & 0.99 & 0.99 \\
% \hline
C\&W ${L_2}$ vs DeepFool ${L_2}$ vs FGSM ${L_2}$ & AlexNet & \textbf{0.66}  & 0.93 & 0.61 & 0.44 \\
& VGG16 & \textbf{0.78}  & 0.91 & 0.80 & 0.63 \\
& ResNet50 & \textbf{0.73} & 0.99 & 0.65  & 0.56 \\
\hline
C\&W ${L_2}$ vs DeepFool ${L_{\infty}}$ vs FGSM ${L_{\infty}}$ & AlexNet & \textbf{0.99} & 0.99 & 0.99  & 0.99 \\
& VGG16 & \textbf{1.00} & 1.00 & 1.00 & 1.00 \\
& ResNet50 & \textbf{0.99} & 0.99 & 0.99 & 0.99 \\
\hline
\end{tabular}
\caption{Reporting attribution accuracy for cross attack algorithms as per class accuracy per architecture (model) on MNIST. The hyperparameters chosen for each attack maximize adversarial accuracy and minimize perceptibility and are described in Table~\ref{table:adv_datasets}. This table presents the overall accuracy of the attack algorithm attribution under `Attribution Accuracy' followed by the breakdown of per class accuracy.}
\label{table:mnist_attack_attrib}
\end{table*}

\subsection{Model Attribution}
The second question we sought to answer through this framework was whether or not a particular open-source architecture leaves behind a characteristic signal in the perturbed adversarial data. We examined model attribution on a per-attack basis in Table~\ref{table:model_attrib}. We note that each attack algorithm could distinguish the particular model that was used to create that variation of the adversarial dataset\footnote{Note that results displayed only cover one set of hyperparameters outlined in Table~\ref{table:adv_datasets}.}. Figure~\ref{fig:cifar10_attacks} and Figure~\ref{fig:mnist_attacks} display the result of performing attack algorithm and model attribution simultaneously. These results support our hypothesis that attribution, specifically for distinguishing models, is possible. 
%Additionally, the combination of attack algorithm and model attribution presented to this attribution framework allows us to distinguish these factors simultaneously, which may aid in characterizing certain adversaries that repeatedly use particular attacks in such a way.
Additionally, by combining attack and model attribution in the framework, we can distinguish both the attack algorithm and the model simultaneously, which may aid in characterizing certain adversaries that repeatedly use particular attacks in such a way.

\begin{table}[ht!]
% \caption{Wide single-column table in a twocolumn document.}
\centering
\begin{tabular}{ccc}%{c{0.3\linewidth}c{0.3\linewidth}c{0.3\linewidth}}
\multicolumn{3}{c}{\textbf{Adversarial Model Attribution:}} \\
\multicolumn{3}{c}{\textbf{AlexNet vs VGG16 vs ResNet50}} \\
\hline
\textbf{Attack} &  \textbf{CIFAR-10}  & \textbf{MNIST}\\
\textbf{Algorithm} &  \textbf{Accuracy} & \textbf{Accuracy} \\
\hline
C\&W ${L_2}$ & \textbf{0.50} & \textbf{0.99} \\
\hline
DeepFool ${L_2}$ & \textbf{0.60} & \textbf{0.84} \\ 
\hline
DeepFool ${L_{\infty}}$ & \textbf{0.60} & \textbf{1.00} \\ 
\hline
FGSM ${L_2}$ & \textbf{0.70} & \textbf{0.87} \\ 
\hline
FGSM ${L_{\infty}}$ & \textbf{0.83} & \textbf{0.99} \\ 
\hline
\end{tabular}
\caption{Reporting best results for model attribution, distinguishing an AlexNet CNN from a VGG16 and ResNet50 architecture, where each attack algorithm hyperparameters are selected by smallest (most imperceptible) perturbation that maximizes adversarial accuracy. Hyperparameters and norm choices are described for each experiment in Table~\ref{table:adv_datasets}.}
\label{table:model_attrib}
\end{table}

% \begin{figure}
% \centering
% \includegraphics[width=8.5cm]{figures/cw_df_fgsm_l2_3models_cm-2.png}
% \caption{An expansion of the three-class Attack Algorithm Attribution to the nine-class Attack Algorithm (and Model) Attribution. Due to rounding errors all rows may not add up to 1.}
% \label{fig:cifar10_attacks}
% \end{figure}

% \begin{figure}
% \centering
% \includegraphics[width=8.5cm]{figures/cw_df_fgsm_l2_3models_cm.png}
% \caption{An expansion of the three-class Attack Algorithm Attribution to the nine-class Attack Algorithm (and Model) Attribution. Due to rounding errors all rows may not add up to 1.}
% \label{fig:mnist_attacks}
% \end{figure}

    \begin{table*}[ht]
    \centering
\begin{tabularx}{\linewidth}{XX}
\includegraphics[width=\columnwidth]{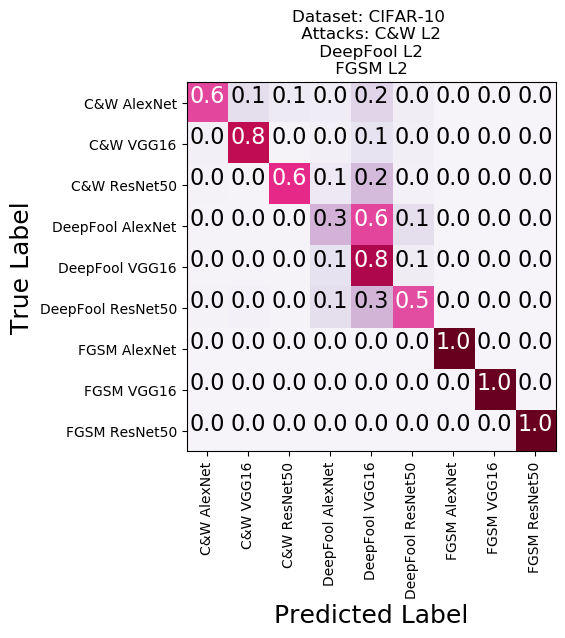}
\captionof{figure}{An expansion of the three-class Attack Algorithm Attribution to the nine-class Attack Algorithm (and Model) Attribution on CIFAR-10. For readability, values below 10\% are not shown, so rows may not sum to 1.}\label{fig:cifar10_attacks}
    &   \includegraphics[width=\columnwidth]{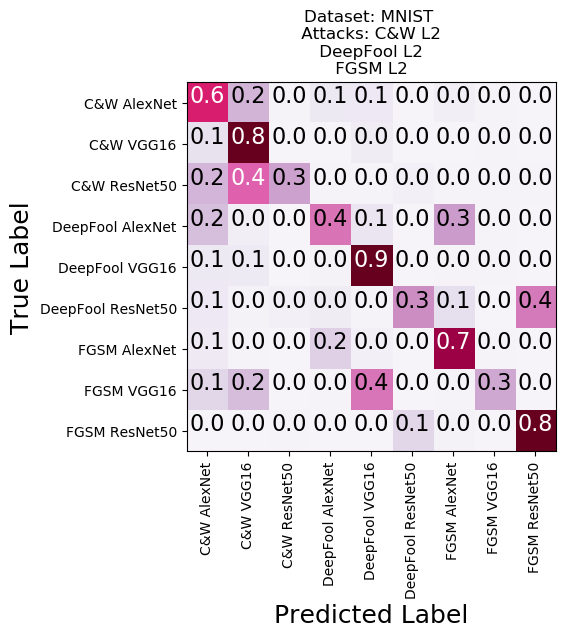}   
        \captionof{figure}{An expansion of the three-class Attack Algorithm Attribution to the nine-class Attack Algorithm (and Model) Attribution on MNIST. For readability, values below 10\% are not shown, so rows may not sum to 1.}\label{fig:mnist_attacks} 
\end{tabularx}
%\caption{A table with figures}
\label{tab:confmatrixfigs}
    \end{table*}%

\subsection{Hyperparameter Attribution}

Our final question we sought to answer through this framework is individual attack algorithm hyperparameter attribution. Each class of attacks has a set of individual hyperparameters that are used to create the attack and are described in Table~\ref{table:adv_datasets}. Attack algorithms are evaluated individually due to the fact that different attack algorithms do not share hyperparameters. An example experimentation for hyperparameter attribution is as follows: using the DeepFool ${L_2}$ attack created with an AlexNet CNN and the hyperparameters from Table~\ref{table:adv_datasets}, hyperparameter attribution aims to distinguish Overshoot values $0.01$, $0.1$, and $1.0$. As shown in Table~\ref{table:cifar_param_attrib}, our attribution classifier fails to be able to find an underlying signal that is attributable to the DeepFool ${L_2}$ attack for each model. This indicates that our framework begins to break down for attacks such as these. Similarly, in Table~\ref{table:mnist_param_attrib}, we see that our framework breaks down for the FGSM ${L_{\infty}}$ attack algorithm for each model on this task. Due to this particular type of attribution failing in some cases while succeeding in others, especially on separate datasets, this type of attribution may not be a reliable distinguishing factor in a real-world scenario. However, more experimentation would need to be done to evaluate whether combinations of attack algorithm or attack model with attack hyperparameters could provide a valuable form of attribution when presented to a system.

It should be noted that as a secondary experiment to verify that our attribution classifier was or was not distinguishing only the `magnitude' of an adversarial example's perturbation, we also conducted an experiment to determine the significance of magnitude of adversarial noise as it pertains to our attribution classifier. Due to the DeepFool and C\&W attacks being optimized, no two examples within each attack dataset have the same magnitude of perturbation, or have the same amount of ``noise" per example. To account for magnitude differences and to ensure that our network is not learning the difference between those magnitudes rather than some underlying signal as we've predicted, we separate or `bin' the datasets into specific ranges using the absolute difference of original to perturbed image. These ranges allowed us to study attribution on the same scale. However, due to class imbalances when separating the adversarial data in this way, conclusions could not be accurately drawn as to whether it had meaningful results. A further study of attack algorithm optimization norm and this ``binning" process could give insights into the magnitude of the perturbation as it pertains to the underlying signal and what our attribution classifier is actually distinguishing. 
 
% The ranges are determined by taking the absolute difference (${L_0}$ norm) between the original image and the perturbed image. The resulting pixels that have been changed are counted and that adversarial example is binned accordingly. The ranges chosen were 0-25\%, 25-50\%, 50-75\%, 75+\% of total pixels in the original perturbed image. We chose four ``bins" because the spread of data into those bins was optimal for training the attribution classifier. Increasing the amount of bins reduced the data samples available per bin, which was not optimal to train a classifier on due to class-imbalances.

%  However, an additional experiment was conducted for hyperparameter attribution in terms of the perturbation magnitude. The results are shown in Table~\ref{table:cifar_magnitude_attrib} in the following section. Similarly, Table~\ref{table:mnist_param_attrib} shows that our attribution classifier fails to distinguish attack hyperparameters for the FGSM ${L_{\infty}}$ attack. This would indicate that the magnitude of the noise the FGSM algorithm produces is similar across different hyperparameters due to the FGSM algorithm not being an optimized attack. 

\begin{table}[ht!]
% \caption{Wide single-column table in a twocolumn document.}
\centering
%\begin{tabular}{c{0.27\linewidth}c{0.14\linewidth}c{0.14\linewidth}c{0.14\linewidth}}
\begin{tabular}{cccc}
\multicolumn{4}{c}{\textbf{CIFAR-10 Adversarial Hyperparameter Attribution}} \\
\hline
\textbf{Attack} &  \textbf{AlexNet} & \textbf{VGG16} & \textbf{ResNet50} \\
\hline
C\&W ${L_2}$ & \textbf{0.66} & \textbf{0.68} & \textbf{0.50}\\
\hline
DeepFool ${L_2}$  & 0.32 & 0.33 &  0.33 \\
\hline
DeepFool ${L_{\infty}}$ & 0.28 & \textbf{0.34} & \textbf{0.36} \\
\hline
FGSM ${L_2}$ & \textbf{0.53} & \textbf{0.48} & \textbf{0.52} \\
\hline
FGSM ${L_{\infty}}$ & \textbf{0.70} & \textbf{0.67} & \textbf{0.70} \\
\hline
\end{tabular}
\caption{Adversarial attack hyperparameter attribution on CIFAR-10.}
\label{table:cifar_param_attrib}
\end{table}

\begin{table}[ht!]
% \caption{Wide single-column table in a twocolumn document.}
\centering
%\begin{tabular}{c{0.27\linewidth}c{0.14\linewidth}c{0.14\linewidth}c{0.14\linewidth}}
\begin{tabular}{cccc}
\multicolumn{4}{c}{\textbf{MNIST Adversarial Hyperparameter Attribution}} \\
\hline
\textbf{Attack} &  \textbf{AlexNet} & \textbf{VGG16} & \textbf{ResNet50} \\
\hline
C\&W ${L_2}$ & \textbf{0.46} & \textbf{0.57} & \textbf{0.49}\\
\hline
DeepFool ${L_2}$  & \textbf{0.49} & \textbf{0.49} &  \textbf{0.49} \\
\hline
DeepFool ${L_{\infty}}$ & \textbf{1.00} & \textbf{0.94} & \textbf{0.91} \\
\hline
FGSM ${L_2}$ & \textbf{0.77} & \textbf{0.42} & \textbf{0.88} \\
\hline
FGSM ${L_{\infty}}$ & 0.32 & 0.32 & 0.32 \\
\hline
\end{tabular}
\caption{Adversarial attack hyperparameter Attribution on MNIST.}
\label{table:mnist_param_attrib}
\end{table}

\begin{figure}
\centering
\includegraphics[width=8cm, height=8cm]{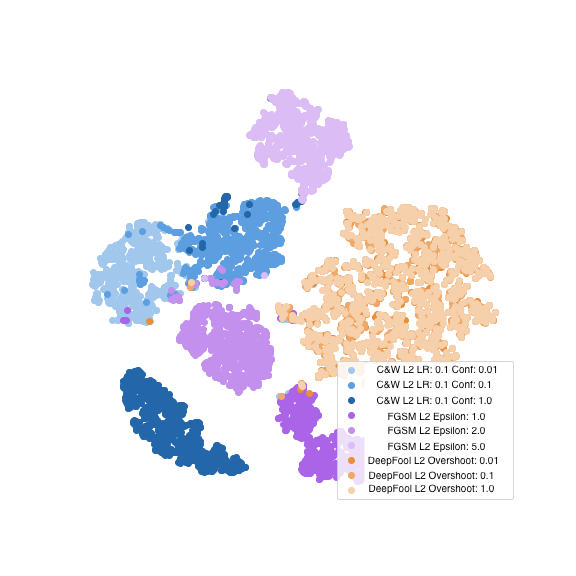}
\caption{t-SNE plot of attack algorithm and hyperparameter attribution as a simultaneous attribution experiment. The three colors represent the attack algorithm class, and shades represent the hyperparameters for each attack algorithm on the CIFAR-10 dataset. }
\label{fig:hyper_param_tsne}
\end{figure}

We also include an additional experiment and visualization shown in Figure~\ref{fig:hyper_param_tsne} to expand our hyperparameter attribution to attack algorithm attribution and hyperparameter attribution simultaneously on the AlexNet model. As we can see from the t-SNE plot, we can clearly distinguish the sub-classes of the attack algorithms C\&W ${L_2}$ and FGSM ${L_2}$, represented by the blue and purple colors, respectively, but fail to make clear distinctions between the sub-classes of the DeepFool ${L_2}$ algorithm represented as orange, which are overlapped and cannot be separated by the attribution classifier. This corresponds to Table~\ref{table:cifar_param_attrib} for the AlexNet model where the DeepFool ${L_2}$ attack algorithm does not perform better than random chance on the hyperparameter attribution experiment. However, we are still able to distinguish the particular attack algorithm as shown in Figure~\ref{fig:hyper_param_tsne} when hyperparameter and attack algorithm are attributed simultaneously. 

\subsubsection{Norm Attribution}

As an additional experiment under the Hyperparameter Attribution, we also sought to answer whether or not particular norms that were used to optimize attack algorithms left behind an underlying signal.
We use individual attack algorithm hyperparameters that maximize adversarial accuracy and minimize perceptibility and are described in Table~\ref{table:adv_datasets}.
Tables~\ref{table:cifar_norm_attrib} and ~\ref{table:mnist_norm_attrib} show that for DeepFool and FGSM, which we optimized using the ${L_2}$ and ${L_{\infty}}$ distances, for each attack algorithm and for each model, the norm used to create the attack is attributable. The break down for each table to show per-class accuracy of the  ${L_{\infty}}$ and ${L_2}$ norms is also presented. 
%This data is important to show that on a per-class and per-model basis neither norm is more robust in being attributable, and that there does exist some underlying signal that aids our attribution classifier in distinguishing norms that optimize the attack algorithm. 

% \begin{table}[thb!]
% \begin{center}
% % CIFAR-10 Adversarial Norm Attribution \\ 
% % \hline
% \raggedright
% \setlength{\tabcolsep}{0pt} % let TeX do the calculations
% \begin{tabular*}{\columnwidth}{
%   @{\extracolsep{\fill}}
%   l % brand
%   *{3}{S[table-format=1.5]}
% }
% \multicolumn{5}{c}{\textbf{CIFAR-10 Adversarial Norm Attribution}} \\
% \hline
% \toprule
% \textbf{Attack}  & \textbf{Accuracy} & \textbf{Model} & \textbf{${L_{\infty}}$} & \textbf{${L_2}$}\\
% \hline
% DeepFool & \textbf{0.60} & AlexNet & 0.60 & 0.60 \\
%  & \textbf{0.60} & VGG16 & 0.60 & 0.60 \\
%  & \textbf{0.60} & ResNet50 & 0.60 & 0.60 \\
% \hline
% FGSM & \textbf{0.85} & AlexNet & 1.00 & 0.90 \\
%  & \textbf{0.90} & VGG16 & 1.00 & 0.80 \\
%   & \textbf{0.70} & ResNet50 & 1.00 & 0.40 \\
% \hline
% \midrule
% \bottomrule
% \end{tabular*}
% \end{center}

% % (1) DeepFool ${L_{\infty}}$
% % (2) DeepFool ${L_2}$
% % (3) FGSM ${L_{\infty}}$
% % (4) FGSM ${L_2}$

% \caption{ We report attribution accuracy for attack algorithms that optimize across multiple norms to create variations of the adversarial datasets. We use individual attack algorithm hyperparameters that maximize adversarial accuracy and minimize perceptibility and are described in Table~\ref{table:adv_datasets}. }
% \label{table:cifar_norm_attrib}
% \end{table}

\begin{table}[ht!]
% \caption{Wide single-column table in a twocolumn document.}
\centering
%\begin{tabular}{c{0.17\linewidth}c{0.17\linewidth}c{0.17\linewidth}c{0.1\linewidth}c{0.1\linewidth}}
\begin{tabular}{ccccc}
\multicolumn{5}{c}{\textbf{CIFAR-10 Adversarial Norm Attribution}} \\
\hline
\textbf{Attack}  & \textbf{Model} & \textbf{Accuracy} & \textbf{${L_{\infty}}$} & \textbf{${L_2}$}\\
\hline
DeepFool & AlexNet & \textbf{0.60} & 0.60 & 0.60 \\
 & VGG16 & \textbf{0.60} & 0.60 & 0.60 \\
 & ResNet50 & \textbf{0.60} & 0.60 & 0.60 \\
\hline
FGSM & AlexNet & \textbf{0.85} & 1.00 & 0.90 \\
 & VGG16 & \textbf{0.90} & 1.00 & 0.80 \\
  & ResNet50 & \textbf{0.70} & 1.00 & 0.40 \\
\hline
\end{tabular}
\caption{Attribution and per-class accuracy for attack algorithms with different $L_p$ norm constraints on CIFAR-10.}
\label{table:cifar_norm_attrib}
\end{table}

\begin{table}[ht!]
% \caption{Wide single-column table in a twocolumn document.}
\centering
%\begin{tabular}{c{0.17\linewidth}c{0.17\linewidth}c{0.17\linewidth}c{0.1\linewidth}c{0.1\linewidth}}
\begin{tabular}{ccccc}
\multicolumn{5}{c}{\textbf{MNIST Adversarial Norm Attribution}} \\
\hline
\textbf{Attack}  & \textbf{Model} & \textbf{Accuracy} & \ \textbf{${L_{\infty}}$} & \textbf{${L_2}$}\\
\hline
DeepFool & AlexNet & \textbf{0.99} & 0.99 & 0.99 \\
 & VGG16 & \textbf{0.99} & 0.99 & 0.99 \\
& ResNet50 & \textbf{0.99} & 0.99 & 0.99 \\
\hline
FGSM & AlexNet & \textbf{1.00} & 1.00 & 1.00 \\
 & VGG16 & \textbf{1.00} & 1.00 &  1.00 \\
  & ResNet50 & \textbf{0.99} & 0.99 & 0.99 \\
\hline
\end{tabular}
\caption{Attribution and per-class accuracy for attack algorithms with different $L_p$ norm constraints on MNIST.}
\label{table:mnist_norm_attrib}
\end{table}

\section{Discussion of Results}
\label{sec:discussion}

% Our experimentation led to a vast number of results that helped to determine where attribution could be further improved within this framework and where there were clear successes. Some of the main results that we would like to focus on are the ability to perform model attribution across different attacks as well as across different datasets. Additionally, we would like to focus on our results for attribution between different optimization norms within the same larger attack class. The class of experiments for Hyperparameter Attribution were less indicative of attribution being possible, as some attack algorithms were not able to distinguish the subtle changes in optimization parameters within its own class. 

Our results from both the CIFAR-10 and MNIST datasets show that attribution, based upon our three main questions, is in fact possible through a very basic experimental setup. To highlight a few experiments of this work that indicate promising areas of attribution for real-world uses, we first examine the attack algorithm attribution and model attribution. Both of these tasks were not only robust in attributing attacks to their respective algorithm, but they were also robust to distinguishing underlying signals within their own attack algorithm. This indicates that with or without the presence of additional attacks inputted into a system, we can identify an algorithm or even a particular model architecture that an adversary may have used and modified for their own needs. We also show that the combination of attack algorithm and model presented to our framework can also be recovered simultaneously with high fidelity, which could aid in identifying bad actors that introduce attacks in such a way. 

However, it is important to note that model attribution may not be the most optimal choice of attribution factor for real-world use-cases. This is due to number of factors including adversaries ensembling models to train adversarial data, in which case our framework would most likely break down due to underlying signals from multiple models being present within a single adversarial example. Additionally, model attribution may be impossible if the adversary gains access to our own attribution model and is able to train adversarial examples that could thwart our framework for attributing it. Such scenarios would render this type of attribution as significantly less robust as other types like attack algorithm attribution or a combination of attack algorithm and attack hyperparameters, however a further study would need to be made to understand this more and how useful model attribution is as its own distinguishing factor.

Our framework does however begin to break down when attempting to attribute the individual hyperparameters to an attack algorithm for both the CIFAR-10 and MNIST datasets. This may be due to the fact that the individual algorithms are not very sensitive to the range of hyperparameter values used and the resultant attacks are highly similar. In the case of FGSM $L_\infty$, it is simply the magnitude of the perturbation that is being varied and the attribution classifier is unable to distinguish between these datasets. Addtionally there is a clear performance difference between the CIFAR-10 and MNIST results and future work is planned to investigate this discrepancy.

%  To further verify this concept we provide additional experimentation not outlined in our initial investigation to understand magnitude of perturbations and whether or not our attribution classifier is learning the differences between the amount of underlying `noise' present in the perturbation or whether it is in fact learning a distinguishing signal. 

Though our experimentation only allows for known classes to be attributed, we note that further research into unknown classes, or unseen attacks, is necessary to realize this work in a real-world setting. It is intractable to assume a system has been trained on an example of every possible adversarial input into a system, however, for the purpose of our experiments, not having an unknown class allowed us to answer the more basic question of whether or not an underlying signal in the adversarial perturbation is present and can be used to distinguish it from other examples.

% Throughout the experimentation shown in the Results section, attack datasets that have minimal perceptible perturbations that maximize adversarial accuracy are used. However, a more fine-grained study of perceptibility and attribution could be made to study whether or not 

%-------------------------------------------------------------------------
\section{Conclusion}
\label{sec:conclusion}

In this work, we have attempted to answer the question of whether we can discover the attribution of an adversarial attack. We have shown through extensive experimentation 
%through isolating attribution variables, 
that there are underlying signals in the attack algorithm, the model, and the hyperparameters used to optimize attacks that a classifier can recover. We believe this opens a new area of adversarial machine learning of interest to many communities. Not only could this provide a means to defending networks against adversarial attacks, but it could mean providing the ability to uncover the actors and/or tool-chains behind real-world attacks. 

% We hope to continue expanding our experiments to further attacks, datasets, optimization parameters, and new frameworks in future work.

We note that inputs into a system are rarely as large and diverse as an entire adversarial dataset. Therefore, to further constrain our problem to the most realistic experimentation possible, we want to investigate iteratively lowering the sample size of the training set throughout the attribution experimentation. This will help us to understand whether attribution is still possible in a real-world setting where the adversary may only attack a system with a single input. 

Additionally, to further expand our experimentation, we also plan to utilize additional open-source frameworks for generating adversarial data, such as Cleverhans and FoolBox\cite{rauber2017foolbox}. These frameworks and others provide open-source code for generating and fine-tuning adversarial attack methods and we anticipate the possibility of attribution being used to distinguish methods that are produced by these frameworks. We also plan to introduce different types of attacks, such as target object detection models, in addition to classification models, to determine if various types of adversarial attacks can be attributed within this framework.

% Additional experiments need to be conducted to evaluate the framework, the Cleverhans Library, used for these experiments and whether a new framework like Foolbox\cite{rauber2017foolbox}, could leave its own underlying signals for similar attacks. 

% \clearpage % not allowed at AAAI
{\small
\bibliography{attribution}
}

\end{document}